\pdfoutput=1

\documentclass[11pt]{article}

\usepackage[final]{acl}

\usepackage{times}
\usepackage{amssymb}
\usepackage{amsmath}
\usepackage{latexsym}

\usepackage[T1]{fontenc}

\usepackage[utf8]{inputenc}

\usepackage{microtype}

\usepackage{inconsolata}

\usepackage{graphicx}
\usepackage{caption}
\usepackage{subcaption}
\usepackage{booktabs}
\usepackage{multirow}
\usepackage{adjustbox}

\title{Are ELECTRA's Sentence Embeddings Beyond Repair? \\ The Case of Semantic Textual Similarity}

\author{
    Ivan Rep \and David Dukić \and Jan Šnajder \\
    TakeLab, Faculty of Electrical Engineering and Computing, University of Zagreb\\
    \texttt{irep2718@gmail.com} \\ \texttt{\{david.dukic, jan.snajder\}@fer.hr}
}

\begin{document}
\maketitle
\begin{abstract}
While BERT produces high-quality sentence embeddings, its pre-training computational cost is a significant drawback. In contrast, ELECTRA provides a cost-effective pre-training objective and downstream task performance improvements, but worse sentence embeddings. The community tacitly stopped utilizing ELECTRA's sentence embeddings for semantic textual similarity (STS). We notice a significant drop in performance for the ELECTRA discriminator's last layer in comparison to prior layers. We explore this drop and propose a way to repair the embeddings using a novel truncated model fine-tuning (TMFT) method. TMFT improves the Spearman correlation coefficient by over $8$ points while increasing parameter efficiency on the STS Benchmark. We extend our analysis to various model sizes, languages, and two other tasks. Further, we discover the surprising efficacy of ELECTRA's generator model, which performs on par with BERT, using significantly fewer parameters and a substantially smaller embedding size. Finally, we observe boosts by combining TMFT with word similarity or domain adaptive pre-training.
\end{abstract}

\section{Introduction}

Pre-trained language models (PLMs) have been a staple in NLP for years, leveraging self-supervised objectives to improve representations for downstream tasks. BERT \citep{devlin-etal-2019-bert}, one of the most widely used PLMs, uses a masked language modeling (MLM) objective for pre-training. The main drawbacks of MLM are the substantial compute cost due to the low masking rate, and the gap between the pre-training task and downstream tasks. To address these issues, \citet{clark2020electra} introduce ELECTRA, which substitutes MLM with replaced token detection (RTD), achieving the same results as BERT but with four times less compute. RTD uses a generator model that corrupts the input, while a discriminator model distinguishes between corrupted and original tokens.


\begin{figure}[t]
    \includegraphics[width=0.5\textwidth]{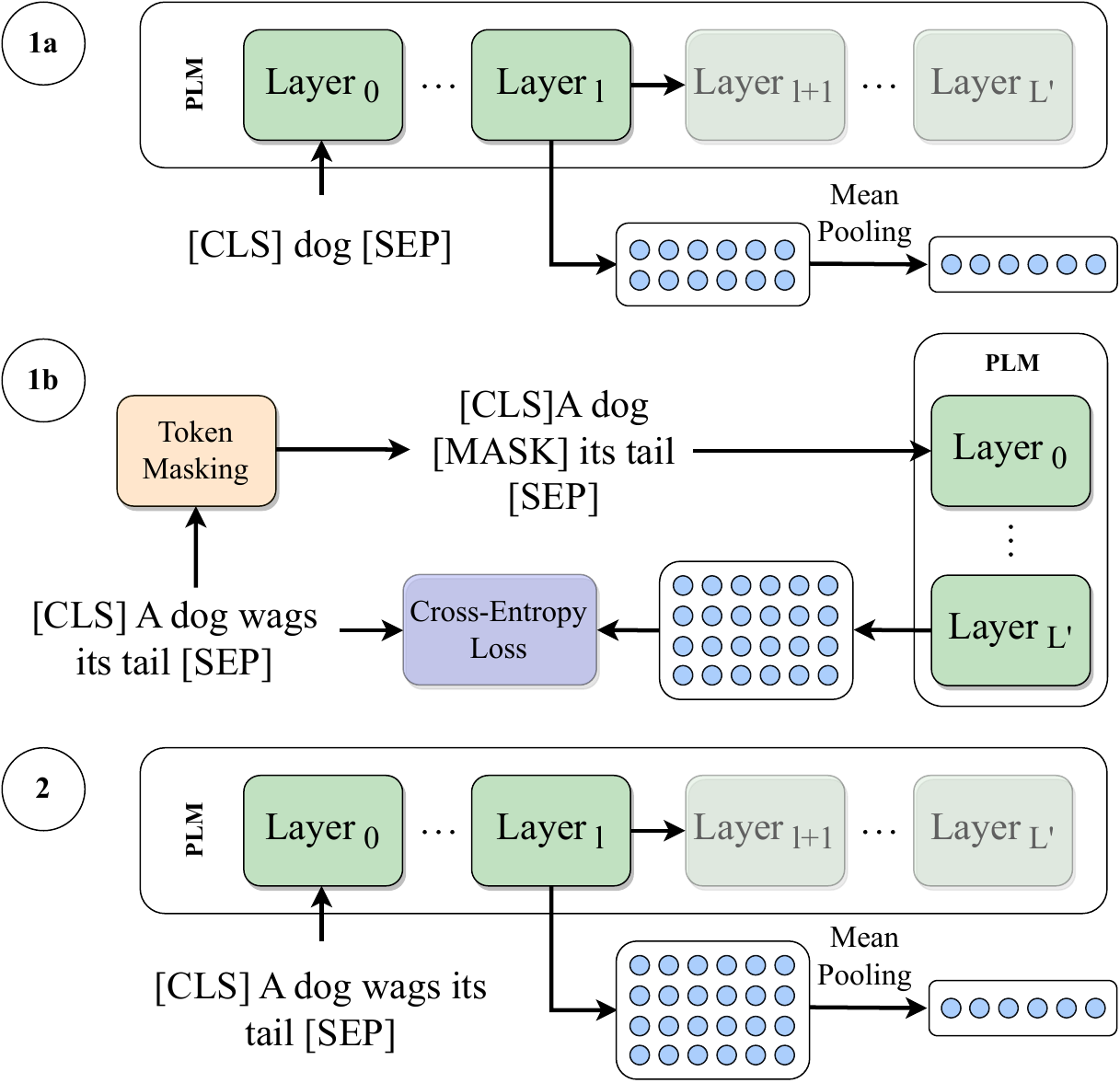}
    \caption{A method for improving sentence embeddings with (2) TMFT on STS. We apply mean pooling over the embeddings at layer $l$ and fine-tune. One of the combinations can also be added for further improvement: (1a) TMFT on word similarity, or (1b) DAPT using MLM.}
    \label{fig:layer-sentence-embedding}
\end{figure}

Semantic textual similarity (STS) \cite{agirre2013sem} is a foundational NLP task with broad applications. 
STS applications must balance accuracy and embedding size to ensure fast inference.
To this end, \citet{reimers-gurevych-2019-sentence} introduced sentence transformers, a framework based on the transformer bi-encoder architecture. It encodes text representations independently before pooling them, calculating the similarity score through a comparison operation between the two embedded texts.

While some transformer models excel across GLUE \citep{wang-etal-2018-glue} tasks, their sentence embedding quality may vary. Sentence transformers often use PLMs pre-trained with a language modeling objective (\citealp{song2020mpnet}; \citealp{raffel2020exploring}; \citealp{liu2019roberta}). In contrast, RTD pre-trained models demonstrate subpar performance in a bi-encoder setting \citep{reimers-hf-talk}, but pre-training such a model is more cost-effective. It is unclear whether this low performance is due to pre-training and whether the embeddings can be improved.

To address this gap, we present a layer-wise study of ELECTRA's sentence embeddings. We examine the performance discrepancy between BERT and ELECTRA's discriminator across various model sizes, languages, and tasks. We hypothesize the last few layers of the discriminator are too specialized for the pre-training task. Following this, we present truncated model fine-tuning (TMFT), shown in Figure~\ref{fig:layer-sentence-embedding}, which utilizes the transformer up to layer $l$, followed by pooling and fine-tuning the whole truncated model. We show that the discriminator suffers from a drop in STS performance when fine-tuning the final layer's embeddings and that this drop is consistent across all inspected model sizes and languages. Although we focus on STS, we expose a similar trend for two other tasks. When applying TMFT on STS Benchmark (STSB) \citep{cer-etal-2017-semeval}, the embeddings significantly outperform the ones from the ELECTRA discriminator's last layer. Moreover, TMFT-obtained embeddings outperform the ones of BERT fine-tuned on STSB up to but not including the eighth layer. Further, we uncover the efficacy of the ELECTRA generator model
, performing on par with BERT while having a significantly smaller embedding size and substantially fewer parameters. Finally, we propose two improvements of the basic TMFT method: prior TMFT on word similarity and prior domain adaptive pre-training (DAPT) using MLM, shown in Figure~\ref{fig:layer-sentence-embedding}.

Our contributions are: (1) a layer-wise analysis of ELECTRA's sentence embeddings for various model sizes, languages, and tasks; (2) a novel TMFT method for repairing ELECTRA's embeddings that substantially improves performance on STS, paraphrase identification, and textual entailment tasks, and exposing the surprising efficacy of the generator model; (3) two additional techniques in combination with TMFT for improving ELECTRA's embeddings even further on STS.  
\footnote{We provide the source code for our work here: \url{https://github.com/ir2718/similarity-embedding-quality}}

\section{Related Work}

\citet{reimers-gurevych-2019-sentence} introduced Siamese networks to transformers, motivating significant research on enhancing sentence embeddings. Fine-tuning on an auxiliary task unrelated to STS has also been explored.
\citet{reimers-gurevych-2019-sentence} first fine-tune on a textual entailment task, followed by fine-tuning on STS. DAPT has also become a widely adopted method, improving performance in downstream tasks \citep{gururangan-etal-2020-dont}.


Similarly, there has been considerable interest in using different layers of a PLM for sentence embeddings. \citet{bommasani-etal-2020-interpreting} assess the layer-wise performance of transformer models and pooling methods on word similarity datasets. \citet{huang-etal-2021-whiteningbert-easy} determine what combinations of hidden states perform the best for unsupervised STS, while \citet{jawahar-etal-2019-bert} extract layer-wise structural characteristics encoded in BERT by probing. Finally, \citet{ethayarajh-2019-contextual} explores layer-wise embedding anisotropy. Two works that resemble ours the most are \citet{hosseini-etal-2023-bert} and \citet{li20242d}. The former improves results by combining layer representations with dynamic programming, while we fine-tune the model from input embeddings up to a specific layer. \citet{li20242d}, developed concurrently with our work, uses model and embedding truncation combined with fine-tuning and a novel loss function. However, this work does not address ELECTRA's performance drop.

\section{Truncated Model Fine-Tuning}

The usual approach for obtaining sentence embeddings is applying a pooling operation over the last layer's embeddings. We use mean pooling as it yields the best results, in line with previous work \citep{reimers-gurevych-2019-sentence}. The TMFT method we propose for repairing ELECTRA's embeddings reduces to taking the $l$-th layer output followed by pooling and fine-tuning on the target task. 

For a sentence $S = (s_1, \hdots, s_N)$ the encoder outputs a tensor $E \in \mathbb{R}^{L' \times N \times d}$, where $L'$ is the number of layers including the input embeddings (which we treat as a layer). We then apply mean pooling $p$ over the $l$-th representation,
$
    p(E, l) = \frac{1}{N} \sum_{n=1}^{N} E_{l, n, :}
$, where $E_{l, n, :}$ is the $d$-dimensional output token embedding of layer $l$ for token at position $n$. We apply this to both sentences, compare them using cosine similarity, and propagate the loss from layer $l$ to the model input.\footnote{Fine-tuning based on the final embedding and using prior embeddings for inference yields unsatisfactory results, and a similar approach is already known \citep{hosseini-etal-2023-bert}.} 


Furthermore, we propose combining one supervised and one self-supervised method with TMFT for performance gains: (1) PLM fine-tuning on the word similarity task or (2) DAPT using MLM. The intuition for the former is that word similarity is crucial for assessing sentence similarity, while the latter builds upon prior evidence that MLM-based PLMs work well in a bi-encoder \citep{reimers-hf-talk}.

\section{Experiments and Results}

We fine-tune each model on the STSB from GLUE. We also run the experiments on machine-translated versions of STSB in Korean, German, and Spanish (cf.~Appendix~\ref{appendix:B}). We choose these based on language-specific dataset and PLM availability. We report the Spearman correlation coefficient, suggested by \citet{reimers-gurevych-2019-sentence}. For word similarity experiments, we use word pairs present in the following datasets: RG-$65$ \citep{10.1145/365628.365657}, WordSim-$353$ \citep{wordsim-353}, SimLex-$999$ \citep{hill-etal-2015-simlex}, and SimVerb-$3500$ \citep{gerz-etal-2016-simverb}, with a random $70$:$15$:$15$ train, validation, and test split. The MLM experiments are conducted on sentences from SNLI \citep{DBLP:journals/corr/BowmanAPM15} and MultiNLI \citep{williams-etal-2018-broad}. We also experiment with paraphrase identification and textual entailment using the Microsoft Research Paraphrase Corpus (MRPC) and Sentences Involving Compositional Knowledge (SICK) textual entailment dataset \citep{marelli-etal-2014-sick}, respectively (cf.~Appendix~\ref{appendix:A}). For STSB, MRPC, and SICK, we use cross-validation splits defined by the dataset authors.
    
We use the following models from HuggingFace Transformers \citep{wolf-etal-2020-transformers}: BERT tiny, mini, small, medium, base, and large. For the ELECTRA discriminator and generator, we use the small, base, and large models. To strengthen our findings on RTD-pre-trained models, we run experiments for the DeBERTaV3 model \citep{DBLP:journals/corr/abs-2111-09543}. For fine-tuning and pre-training, we use the AdamW optimizer \cite{loshchilov2017decoupled}, a learning rate of $2\mathrm{e}{-5}$, and weight decay set to $1\mathrm{e}{-2}$. We apply gradient clipping to a max norm of $1.0$. For fine-tuning on all tasks except word similarity, we use a batch size of $32$ for $10$ epochs. Word similarity fine-tuning uses a batch size of $128$ for $50$ epochs. For DAPT, we use a batch size of $32$ with $8$ gradient accumulation steps for $10$ epochs with $0.15$ masking probability. An exception is DeBERTaV3, with a batch size of $8$ and $32$ gradient accumulation steps due to memory constraints. All reported results are averaged across five seeds. Test set Spearman correlation coefficients on STSB correspond to the model with the highest Spearman correlation coefficient on the validation set. For MRPC and SICK, the test set F1 scores correspond to the models with the classification threshold optimized for the highest validation set F1 score.

\subsection{ELECTRA}

\begin{figure*}[t]
\centering
\begin{subfigure}[t]{0.33\textwidth}
    \centering
    \includegraphics[width=1.0\linewidth]{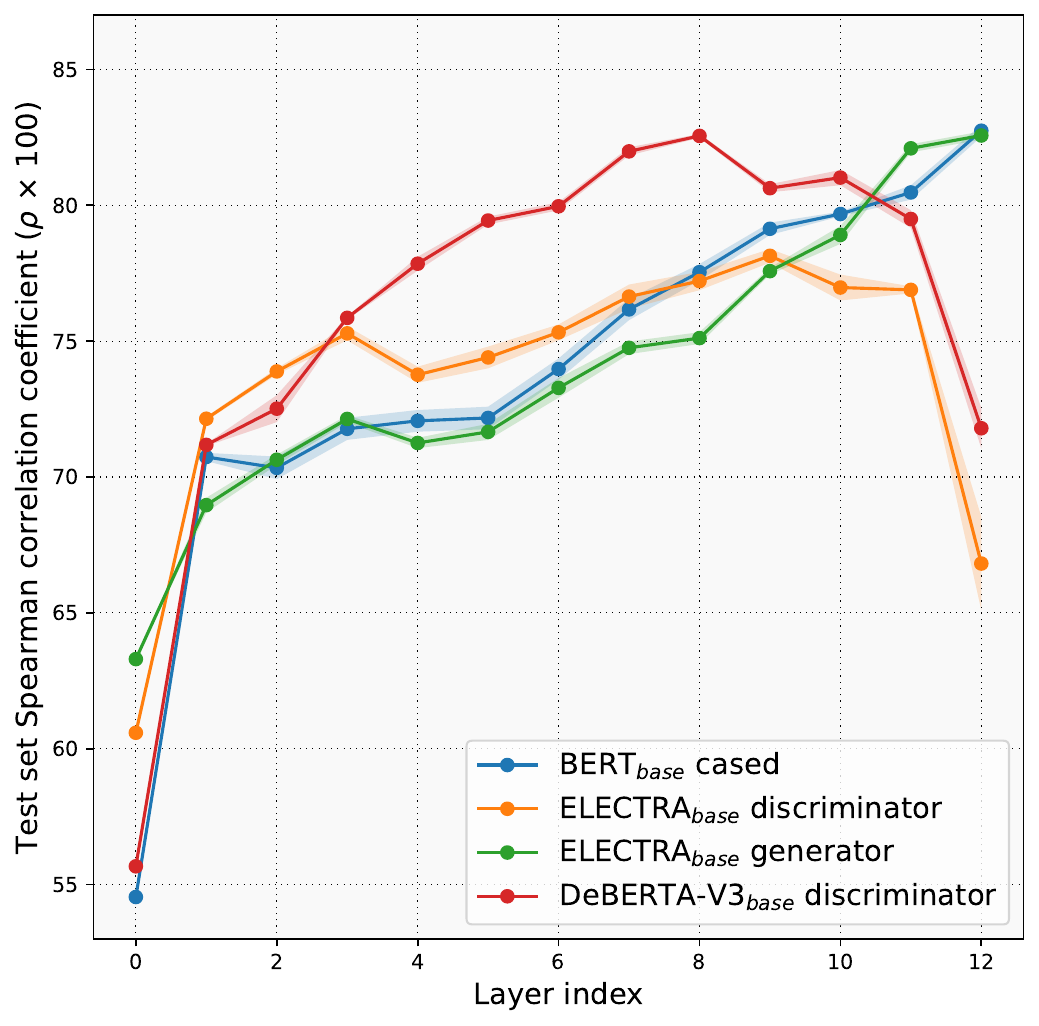}
    \caption{TMFT on STSB}
    \label{fig:comparison-plot}
\end{subfigure}\hfill%
\begin{subfigure}[t]{0.33\textwidth}
    \centering
    \includegraphics[width=0.965\linewidth]{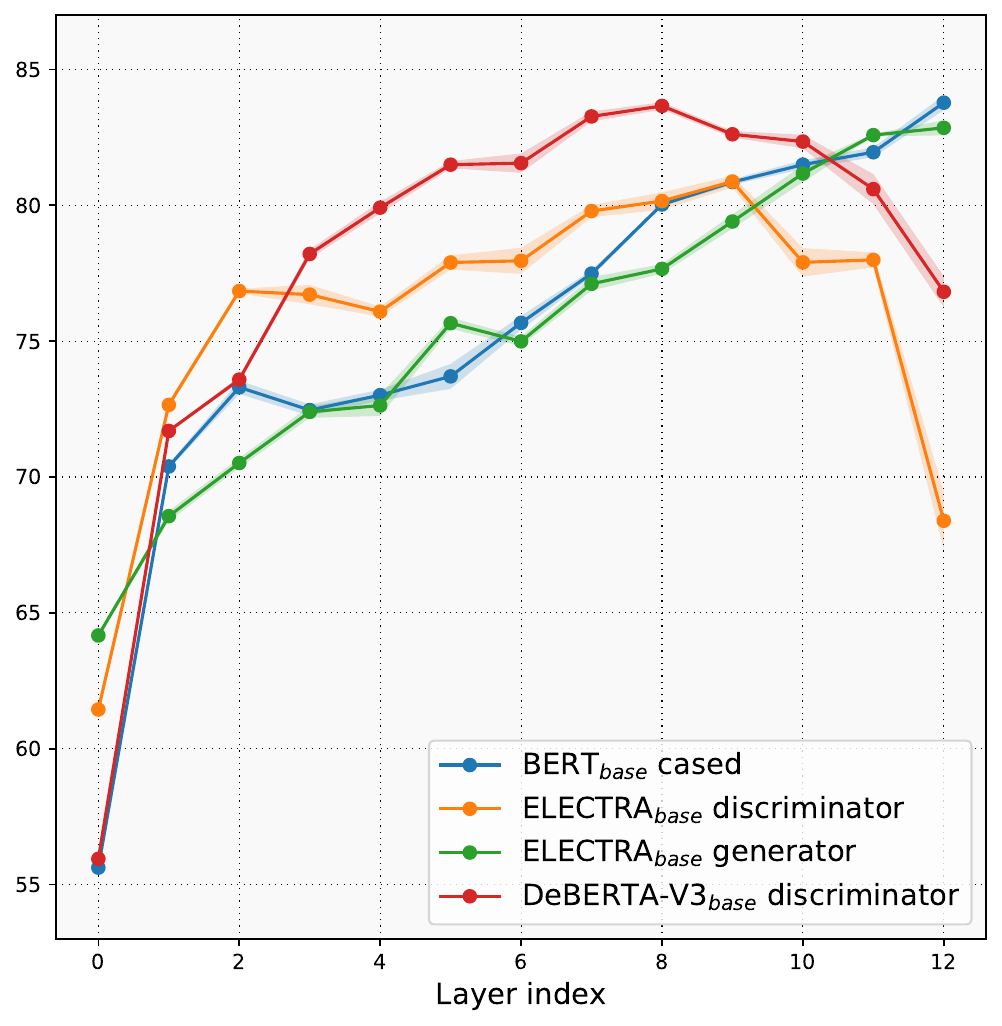}
    \caption{\centering{TMFT on word similarity datasets + TMFT on STSB}}
    \label{fig:comparison-plot-wordsim}
\end{subfigure}\hfill%
\begin{subfigure}[t]{0.33\textwidth}
    \centering
    \includegraphics[width=0.965\linewidth]{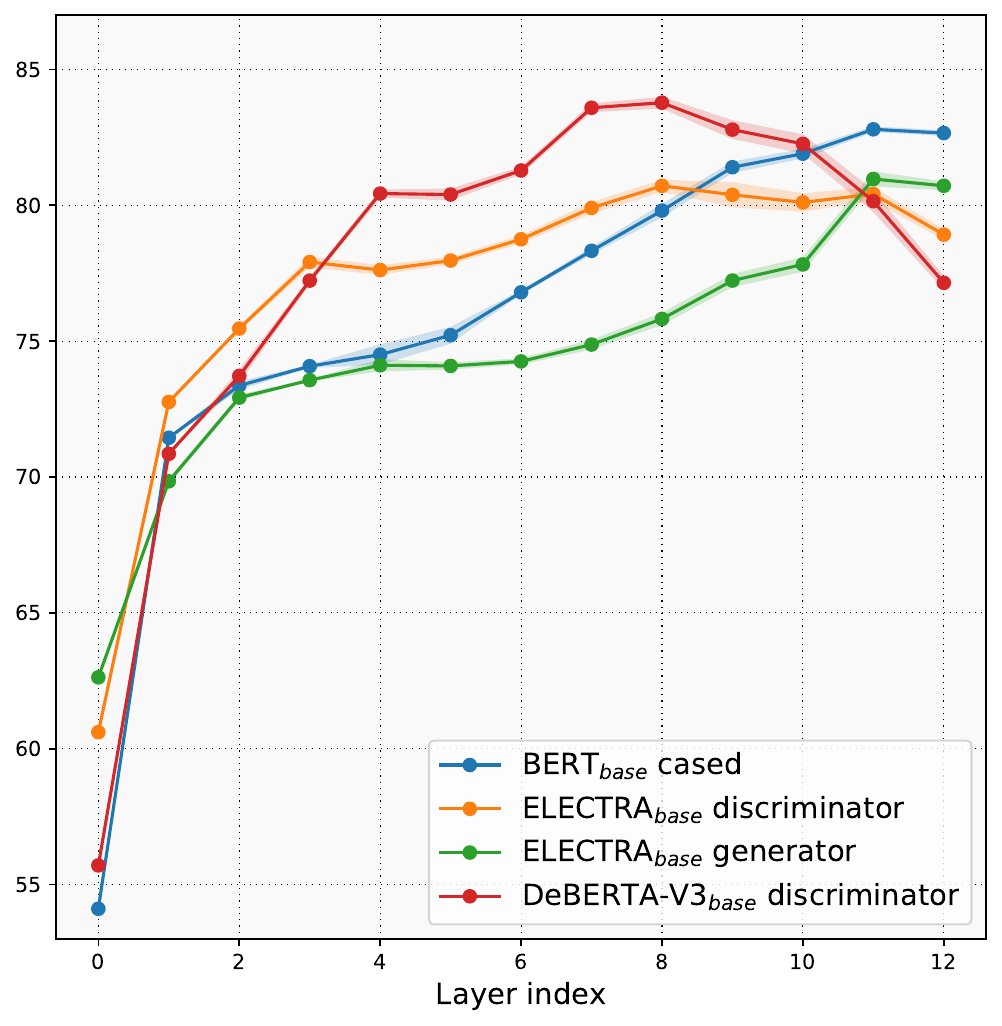}
    \caption{\centering{MLM on NLI datasets + TMFT on STSB}}
    \label{fig:comparison-plot-mlm}     
\end{subfigure}
\caption{Test set Spearman correlation coefficients on STSB using TMFT with and without improvements (shaded area is the standard deviation). Subfigure~\ref{fig:comparison-plot} presents results using TMFT on STSB,~\ref{fig:comparison-plot-wordsim} shows TMFT on STSB with prior TMFT on word similarity, and~\ref{fig:comparison-plot-mlm} depicts TMFT on STSB with prior MLM. More details are in Table~\ref{tab:model-comparison}.}
\label{fig:model-truncation}
\end{figure*}

\begin{table*}[h]
\centering
\begin{adjustbox}{width=\textwidth}
    \begin{tabular}{l|lrcc|lrcc|lrcc}
        \toprule
        {} & \multicolumn{4}{|c|}{\textbf{TMFT STSB}} & \multicolumn{4}{|c|}{\textbf{TMFT WS + TMFT STSB}}  & \multicolumn{4}{|c}{\textbf{DAPT NLI + TMFT STSB}} \\
        \midrule
        \textbf{Model} & \textbf{Layer} & \textbf{Params}  & \textbf{Val} & \textbf{Test} & \textbf{Layer} & \textbf{Params} & \textbf{Val} & \textbf{Test} &  \textbf{Layer} & \textbf{Params}  & \textbf{Val} & \textbf{Test}   \\
        \midrule
        BERT$_{\text{base}}$ & 12 & 107.72M & 86.07/85.98 & \textbf{82.74}/\textbf{83.03}  &  12 & 107.72M & 85.85/85.84 & \textbf{83.77}/\textbf{84.08} &  12 & 107.72M & 85.91/85.71 & 82.66/82.64 \\
        ELECTRA$_{\text{D\:base}}$ & 3 & 45.10M & 82.15/82.20 & 75.29/76.96 & 3 &  45.10M  & 82.66/82.56 & 76.71/77.31 & 7 & 73.45M & 84.07/83.78 & 79.90/79.92 \\
        ELECTRA$_{\text{G\:base}}$ & 12 & 33.31M & \textbf{86.62}/\textbf{86.38} & 82.57/82.50 & 12 & 33.31M & \textbf{86.67}/\textbf{86.39} & 82.85/82.91 & 11 & 32.52M & 85.58/85.22 & 80.97/80.85  \\
        DeBERTaV3$_{\text{base}}$ & 7 & 148.00M & 84.80/84.86 & 81.98/82.44 & 7 & 148.00M & 85.48/85.55 & 83.27/83.31 & 7 & 148.00M & \textbf{85.95}/\textbf{85.88} & \textbf{83.59}/\textbf{83.51}\\
        \bottomrule
    \end{tabular}
\end{adjustbox}
\caption{Comparison of models with the highest validation set Spearman correlation coefficient using TMFT. The reported scores are the test set Spearman and Pearson correlation coefficients. Results with the addition of improvements are included (WS stands for word similarity, NLI stands for natural language inference). Bold values represent the highest values for the used method across all trained models.}
\label{tab:model-comparison}
\end{table*}

Applying ELECTRA to downstream tasks is usually done using the discriminator. We decided not to follow this practice, as ELECTRA's authors do not give convincing reasons for discarding the generator. Hence, we conduct experiments with both models. The generator is similar to BERT, except the generator's input embeddings are tied to the discriminator in pre-training. For comparison, we use BERT as a baseline as it is a standard choice and similar in size to the discriminator.

\begin{figure*}[t!]
\centering
\begin{subfigure}[t]{0.33\textwidth}
    \centering
    \includegraphics[width=1.0\linewidth]{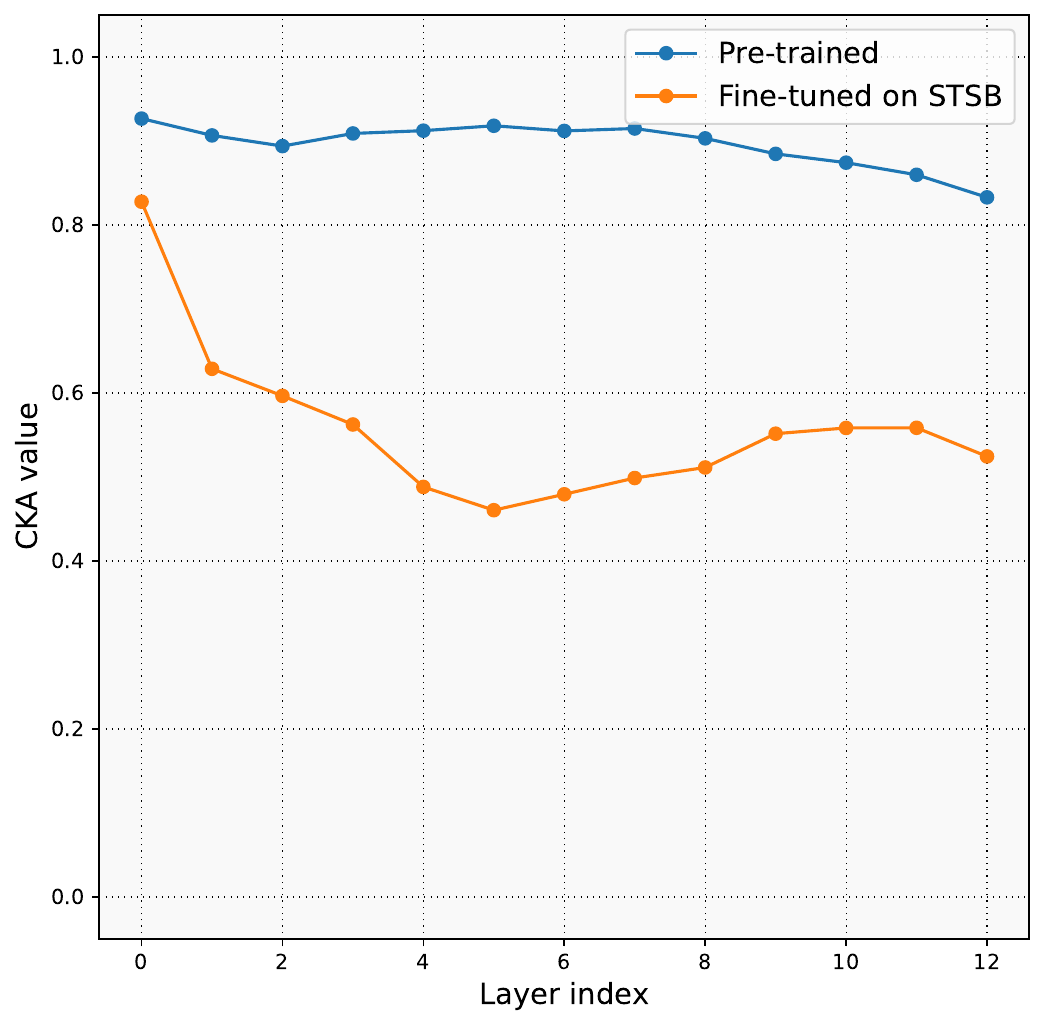}
    \caption{ELECTRA$_{\text{G\:base}}$ and BERT$_{\text{base}}$}
    \label{fig:cka_g_bert}
\end{subfigure}\hfill%
\begin{subfigure}[t]{0.33\textwidth}
    \centering
    \includegraphics[width=0.965\linewidth]{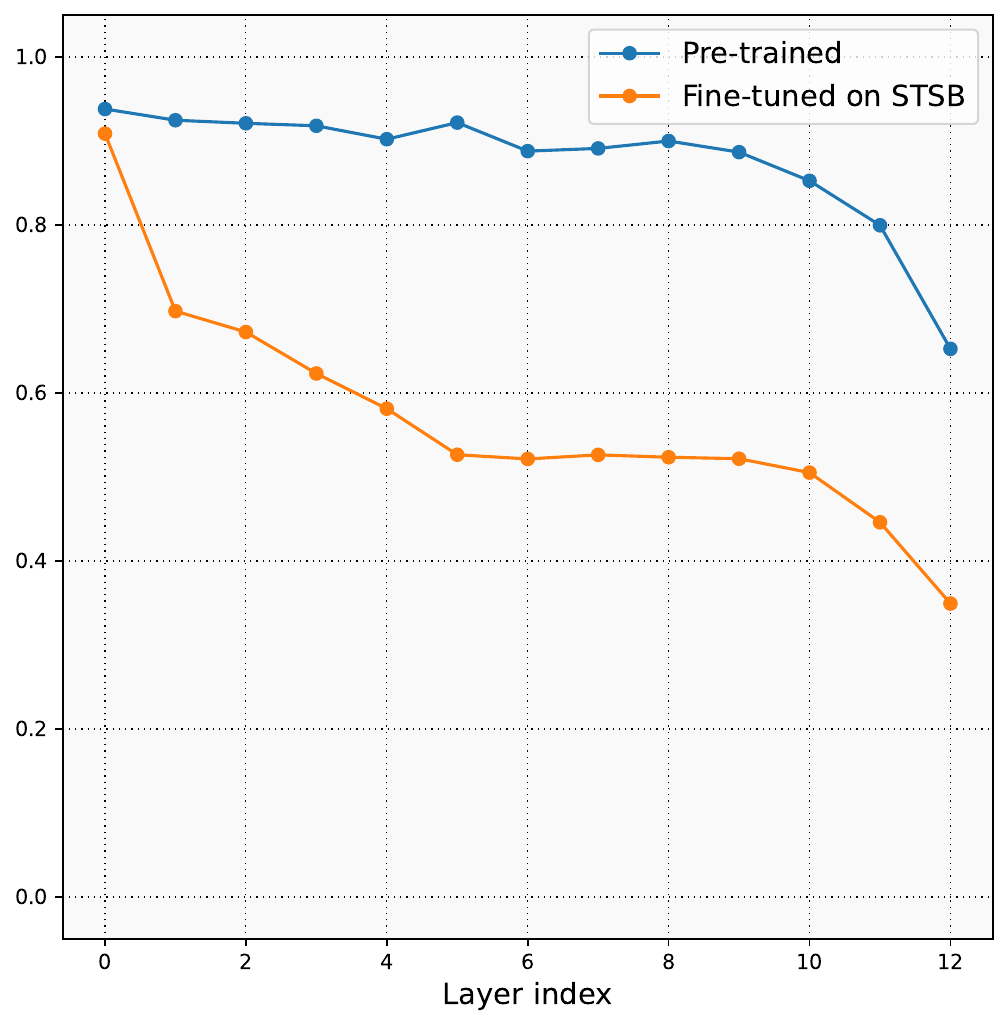}
    \caption{\centering{ELECTRA$_{\text{D\:base}}$  and BERT$_{\text{base}}$}}
    \label{fig:cka_d_bert}
\end{subfigure}\hfill%
\begin{subfigure}[t]{0.33\textwidth}
    \centering
    \includegraphics[width=0.965\linewidth]{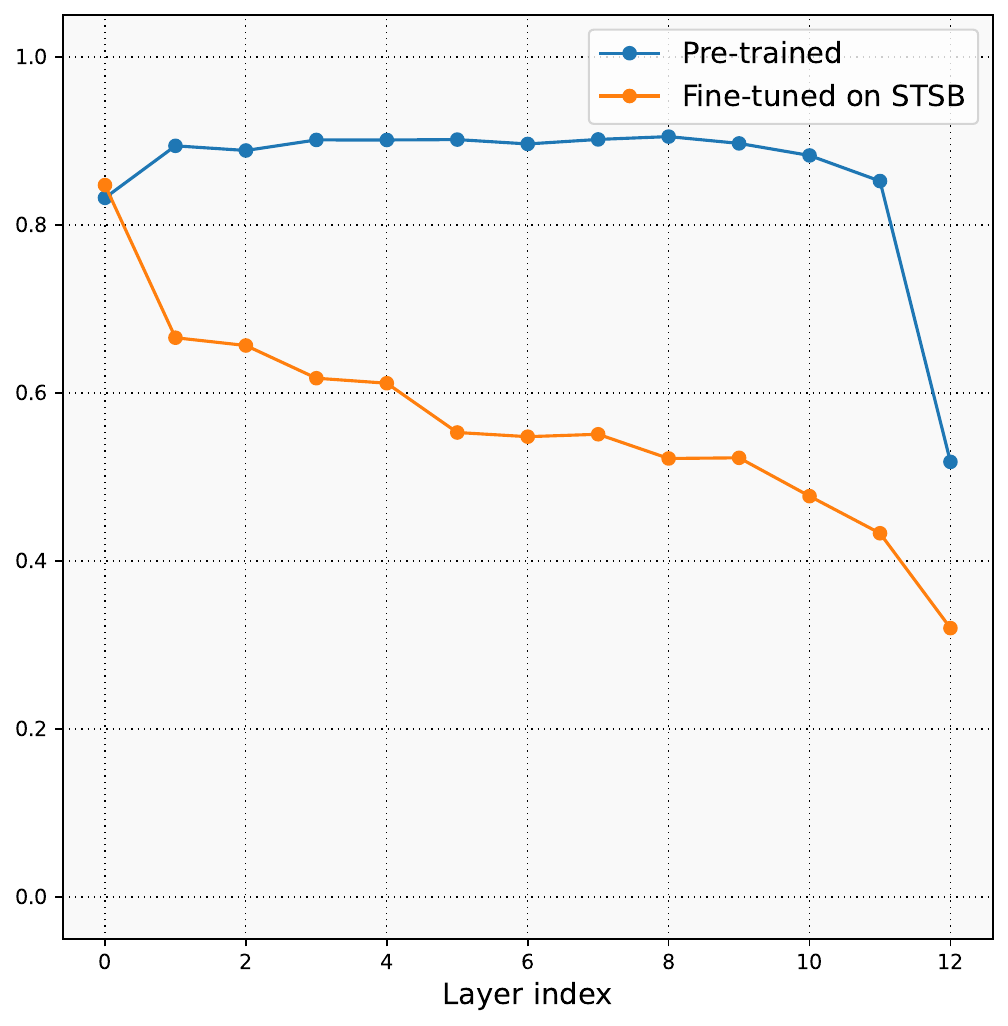}
    \caption{\centering{DeBERTaV3$_{\text{D\:base}}$  and BERT$_{\text{base}}$}}
    \label{fig:cka_deberta_bert}     
\end{subfigure}
\caption{The result of applying CKA on the hidden layer representations of the STSB test set at a layer with a certain index. Subfigure~\ref{fig:cka_g_bert} presents the comparison between ELECTRA generator and BERT, subfigure~\ref{fig:cka_d_bert} the comparison between ELECTRA discriminator and BERT, and subfigure~\ref{fig:cka_deberta_bert} the comparison between DeBERTaV3 discriminator and BERT.}
\label{fig:representation-analysis}
\end{figure*}

Figure~\ref{fig:comparison-plot} shows  test set Spearman correlation coefficients for TMFT applied to ELECTRA$_{\text{base}}$ discriminator, ELECTRA$_{\text{base}}$ generator, and BERT$_{\text{base}}$. BERT shows a trend where the Spearman correlation coefficient roughly increases as the index of the fine-tuned layer embedding increases. The same trend is present for the generator. ELECTRA$_{\text{base}}$ generator with $33.31$M parameters maintains comparable performance to BERT$_{\text{base}}$ with $107.72$M parameters on all tasks (cf.~Table~\ref{tab:model-comparison}, Figure~\ref{fig:mrpc}, and Figure~\ref{fig:sick-e} in Appendix~\ref{appendix:A}). This finding is consistent across all inspected generator and BERT sizes (cf.~Table~\ref{tab:pareto-table} in Appendix~\ref{appendix:D}). The discriminator shows a different trend, gradually increasing until the ninth layer, after which performance drops sharply. We attribute this drop to the RTD task, also suggested by Centered Kernel Alignment (CKA) representation similarity analysis, which shows that CKA values between discriminator models and MLMs drop in the final layers even before fine-tuning (cf.~Figure~\ref{fig:representation-analysis}). A similar performance drop occurs for Korean, German, and Spanish (cf.~Appendix~\ref{appendix:B}), all considered discriminator model sizes (cf.~Figure~\ref{fig:discriminator}), the paraphrase identification task (cf.~Figure~\ref{fig:mrpc}), and the DeBERTaV3 discriminator (cf.~Figure~\ref{fig:comparison-plot}). The increase in the ELECTRA discriminator test Spearman correlation coefficients between the ninth ($87.63$M parameters) and last state ($108.89$M parameters) is $11.32$ points. The best-performing state on the validation set is the third state ($45.10$M parameters), and the increase for the third state compared to the last state on the test set is $8.36$. Furthermore, the discriminator's input embeddings outperform BERT, while its output embeddings surpass BERT's up to, but not including, the eighth layer. Across all models, the largest increase between consecutive layers is between the 0th and 1st layers, likely due to self-attention.

\subsection{Further Improvements}

Our first proposed improvement is TMFT on word similarity before TMFT on sentence similarity. The downside of this method is that it requires data labeled for word similarity. We consider only using the same layer for both fine-tuning procedures. Figure~\ref{fig:comparison-plot-wordsim} gives the test set Spearman correlation coefficients on STSB for the proposed improvement. We observe BERT and the discriminator both benefit from this method, with ELECTRA outperforming BERT up to but not including the tenth layer.

Our second improvement is DAPT using MLM before TMFT on STS, regardless of the model type. We only consider pre-training using the last layer's output. Figure~\ref{fig:comparison-plot-mlm} presents the test set Spearman correlation coefficients on the STSB dataset. With improvements included, the performance of almost all representations improved across all models, and the drop in the final layers of the discriminator is diminished (cf.~Figure~\ref{fig:comparison-plot-mlm} and Table~\ref{tab:model-comparison}).

\begin{figure}[h!]
\centering
    \includegraphics[width=1\linewidth]{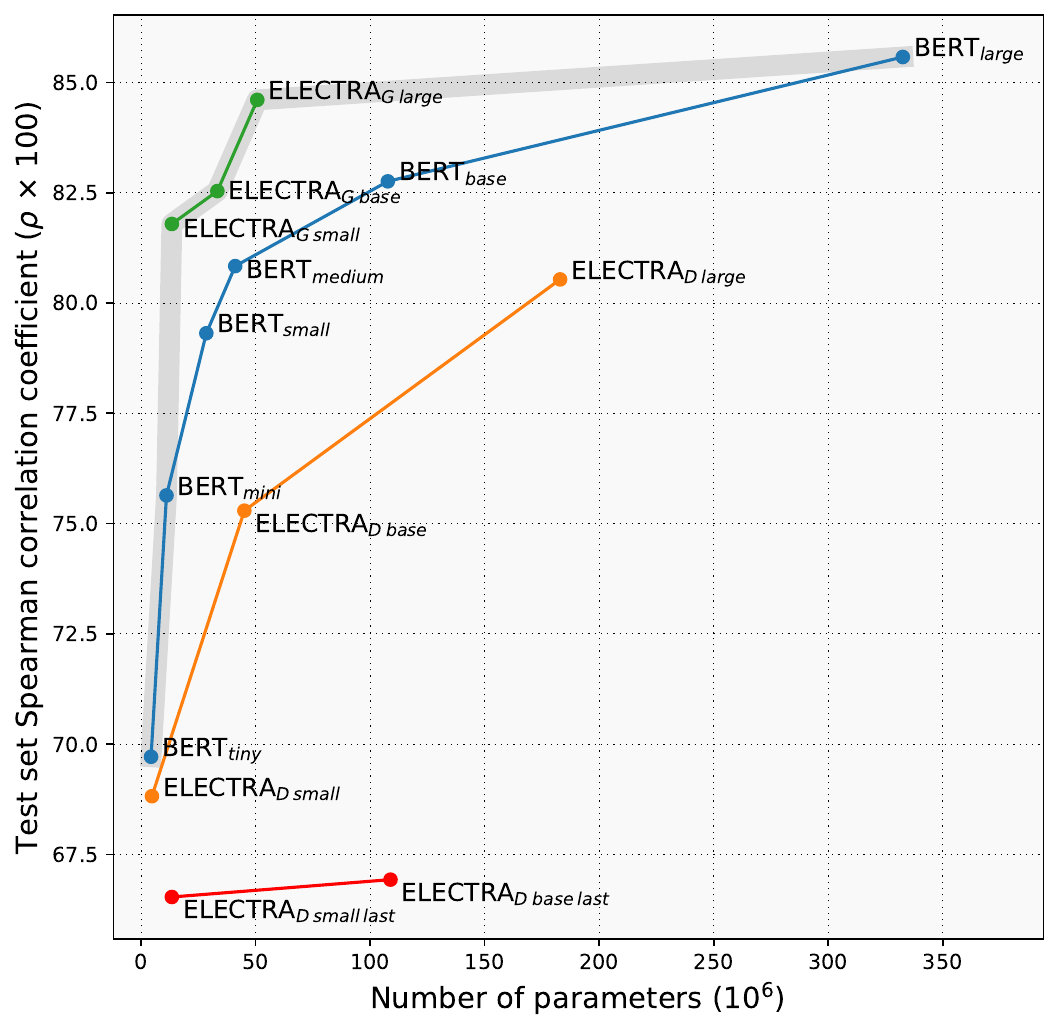}
    \caption{Comparison of the number of parameters of the model and the test set Spearman correlation coefficients. The shown models have the highest validation Spearman correlation coefficient value. The figure also includes the last layer representations that do not correspond to the highest validation Spearman correlation coefficient. ELECTRA$_{\text{large}}$ discriminator is excluded as its value is too small ($25.84$). The gray line indicates the Pareto front. For detailed test set Spearman correlation coefficient values, refer to Table~\ref{tab:pareto-table} in Appendix~\ref{appendix:D}.}
    \label{fig:pareto}
\end{figure}

\subsection{Parameter-Performance Trade-off}

Finally, we investigate the parameter-performance trade-off. The results show the number of parameters the model has when using the best-performing representation on the validation set and the corresponding test set Spearman correlation coefficients (cf.~Figure~\ref{fig:pareto} and Table~\ref{tab:pareto-table}). We consider only TMFT without improvements. Considering the test set scores and the number of parameters, the best models are BERT (tiny, mini, large) and ELECTRA generator (small, base, large).
The difference in the number of parameters and layers suggests the depth of a transformer is essential for retaining performance on STS. Figure~\ref{fig:pareto} also demonstrates an improvement in the number of parameters, and in the performance when comparing the last and best representation of the discriminator.

\subsection{Performance Drop Analysis}
To explain the drop in the Spearman correlation coefficient, we conduct an analysis using CKA \citep{kornblith2019similarity}. We apply the CKA to the STSB test set representations obtained from the hidden states of two models with the same index. Since the time complexity of the TMFT method depends on the dataset size and the number of hidden states, it is beneficial to know about the presence of the possible drop prior to fine-tuning. An additional benefit of this method is that it works with pre-trained models, saving time and resources. We observe a pattern where the drop in CKA values in the final layers is a necessary condition for the drop in Spearman correlation coefficient values in the final layers. The comparison between the ELECTRA discriminator and BERT (Figure~\ref{fig:cka_d_bert}) exhibits the same trend as the comparison between the DeBERTaV3 discriminator and BERT (Figure~\ref{fig:cka_deberta_bert}), which is a sharp decline in CKA values in the final few layers. The comparison between the ELECTRA generator and BERT (Figure~\ref{fig:cka_g_bert}) does not exhibit the CKA value drop. All combinations of inspected models after fine-tuning show similar trends compared to their pre-trained counterparts.
The drop could be attributed to the model architecture, pre-training data, pre-training method, fine-tuning data, or hyperparameter choice. To control for these confounders, we apply TMFT to randomly initialized models. We found that all models exhibit the same behavior, suggesting the architecture and tokenizer choice are not to blame for the performance drop (cf.~Figure~\ref{fig:random} in Appendix~\ref{appendix:E}). 

\section{Conclusion}

We analyze ELECTRA's sentence embeddings for STS and two other tasks, comparing them to a BERT baseline on the STSB dataset. Our proposed truncated model fine-tuning method significantly improves the discriminator. The generator model matches BERT's performance while significantly reducing the number of parameters and producing smaller embeddings. 

\section{Limitations}

Our study reveals substantial differences in model performance on some tasks, but sentence embedding models can also be used for information retrieval. We limited ourselves to one dataset for STS, paraphrase identification, and entailment, four datasets for word similarity, and two datasets for MLM, due to computation restrictions. Experiments in Korean, German, and Spanish were exclusively conducted using the truncated model fine-tuning method. Another limitation of our work is the way we apply word similarity fine-tuning prior to truncated model fine-tuning on STS. We only consider fine-tuning the same layer's embeddings for both procedures. Furthermore, in MLM, prior to truncated model fine-tuning on STS, we only use the last layer's embeddings, as opposed to using previous layers. For all our experiments, we fixed the hyperparameters. Finally, our study covers only two families of models in detail: BERT and ELECTRA.

\bibliography{anthology,custom}


\appendix

\newpage

\section{Performance on Additional Tasks}
\label{appendix:A}

\begin{figure}[h!]
    \centering
    \includegraphics[width=0.8\linewidth]{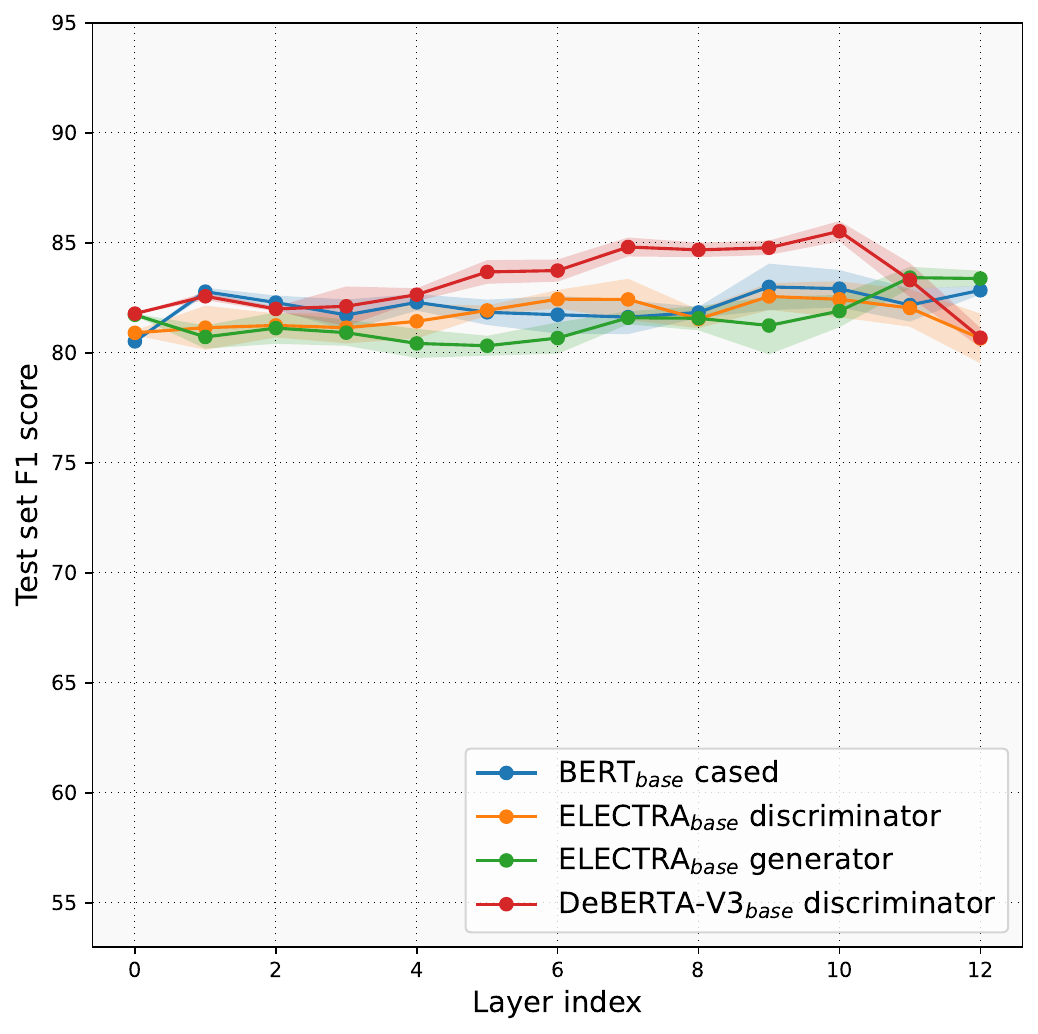}
    \caption{Test set F1 scores on the MRPC dataset across all layers.}
    \label{fig:mrpc}
\end{figure}

Figure~\ref{fig:mrpc} represents the F1 test scores on the MRPC dataset. We use a bi-encoder optimized using a binary cross entropy loss. All other fine-tuning hyperparameters are the same as in the case of STS. ELECTRA discriminator and DeBERTaV3 show the same performance drop in the final layers. ELECTRA achieves an F1 score of $80.66$ for the final hidden state, while the F1 for the ninth hidden state is $82.57$. For DeBERTaV3 the difference is even greater. It achieves an F1 score of  $80.69$ for the final hidden state, while the tenth hidden state achieves $85.53$. BERT and ELECTRA generator do not show a similar trend.

\vfill

\begin{figure}[h!]
    \centering
    \includegraphics[width=0.8\linewidth]{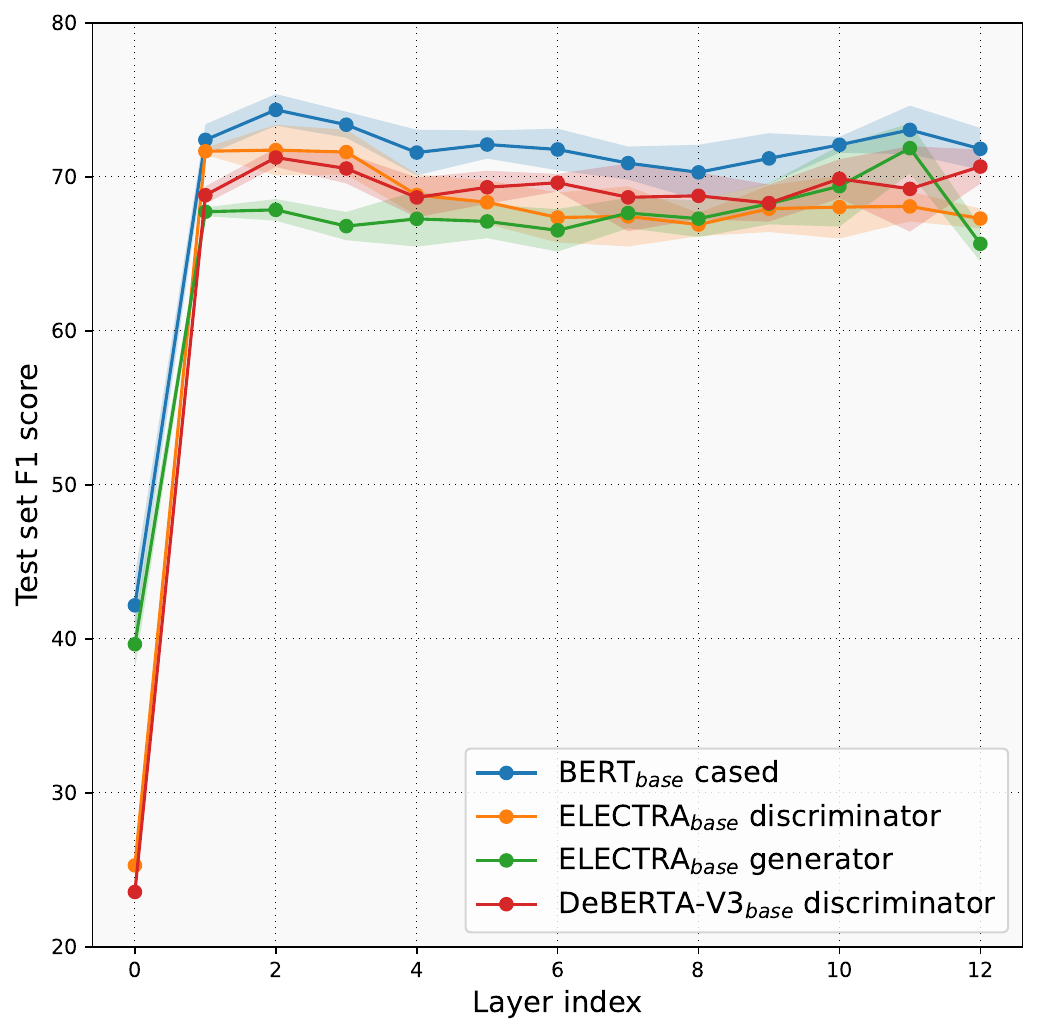}
    \caption{Test set F1 scores on the SICK entailment dataset across all layers.}
    \label{fig:sick-e}
\end{figure}

Figure~\ref{fig:sick-e} represents the F1 test scores on the SICK entailment dataset. We use a bi-encoder to encode the first sentence $u$, the second sentence $v$, and concatenate its absolute difference $|u-v|$ following \citep{reimers-gurevych-2019-sentence}. Finally, a classification head is used to calculate the output probabilities. The model is optimized using a cross entropy loss. All other fine-tuning hyperparameters are the same as in the case of STS. The results show slight performance drops for ELECTRA discriminator in the final layers, while DeBERTaV3 shows an upward trend in the final layers. BERT shows a slight performance drop in the final layer, while the ELECTRA generator exhibits a very large drop. The F1 value for the final layer is $65.66$, while the previous layer achieves $71.87$.

\vfill

\section{Performance on STSB in Different Languages}
\label{appendix:B}

\begin{figure}[h!]
    \centering
    \includegraphics[width=0.8\linewidth]{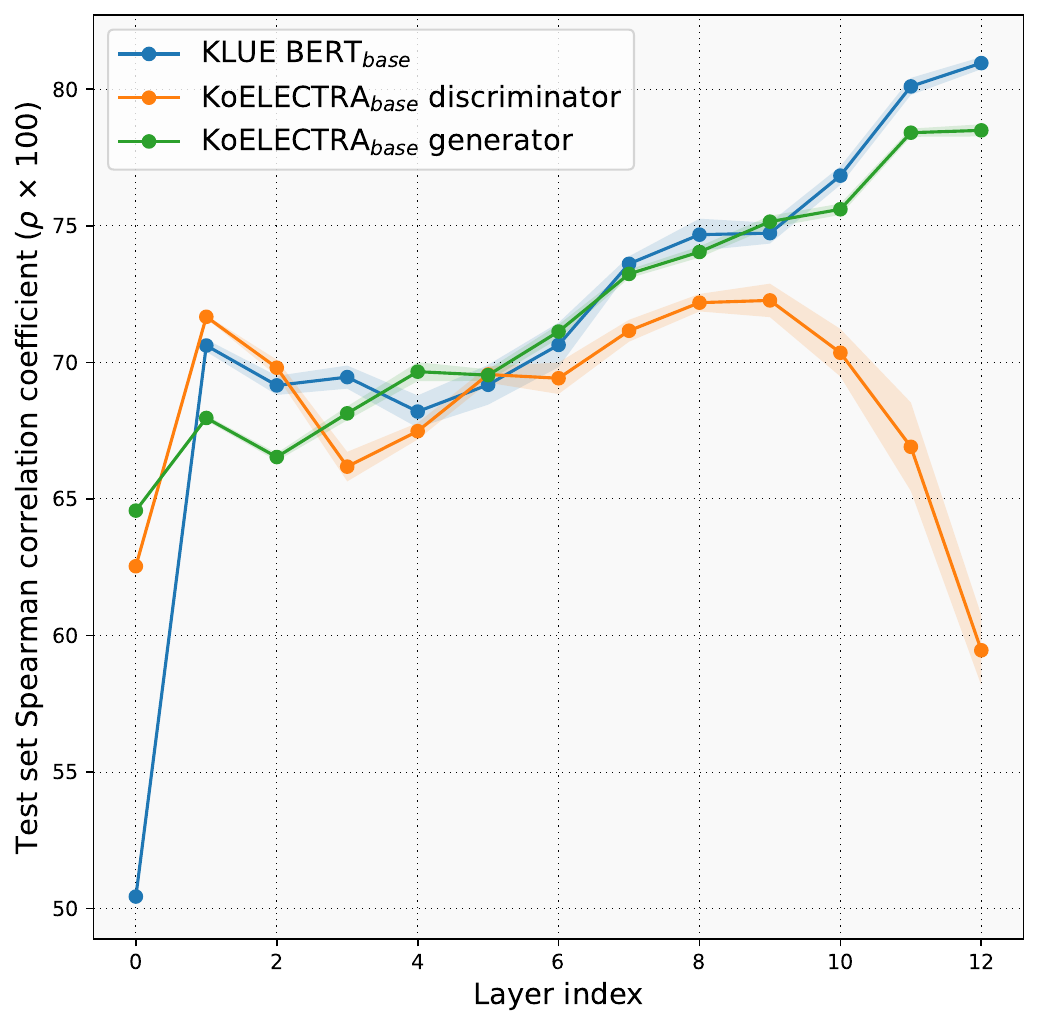}
    \caption{Test set Spearman correlation coefficients on Korean STSB across all layers.}
    \label{fig:korean}
\end{figure}

\begin{figure}[h!]
    \centering
    \includegraphics[width=0.8\linewidth]{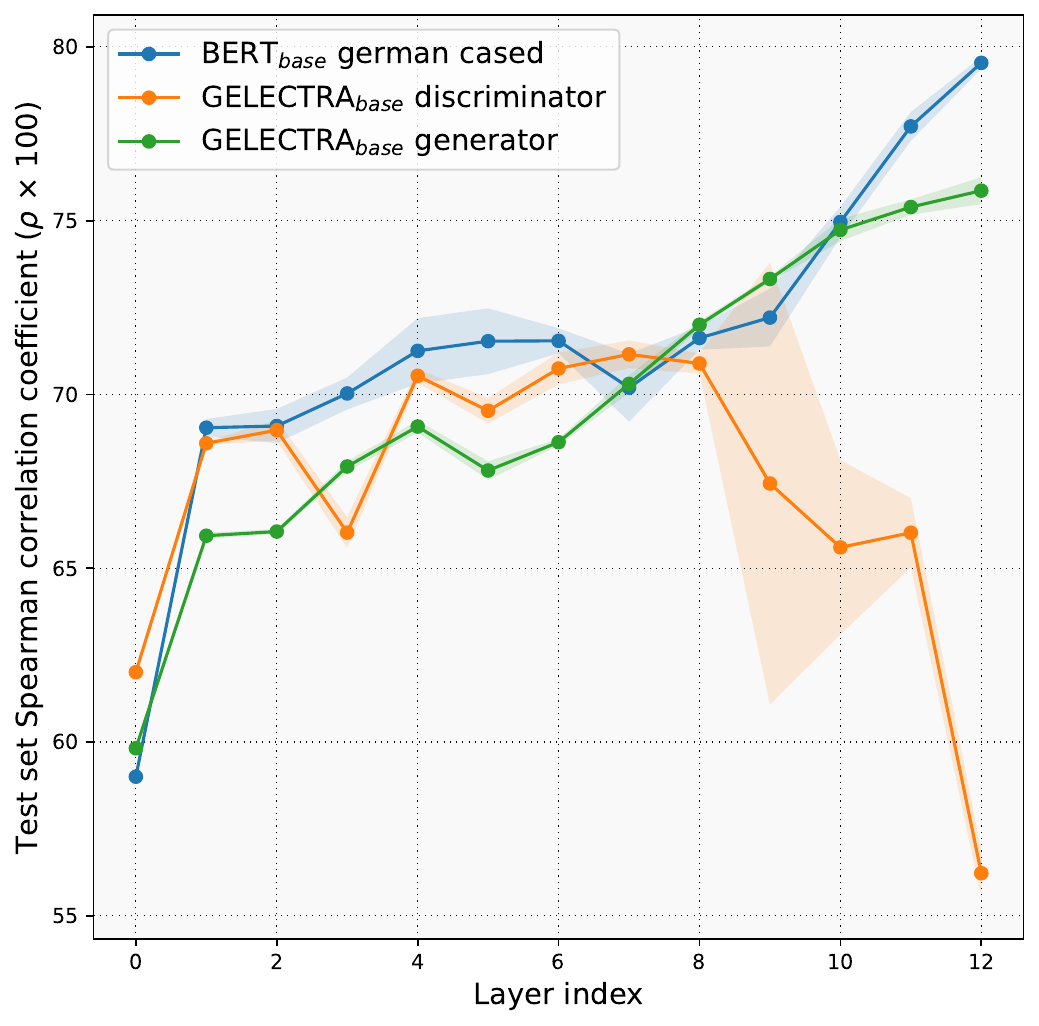}
    \caption{Test set Spearman correlation coefficients on German STSB across all layers.}
    \label{fig:german}
\end{figure}

\begin{figure}[h!]
    \centering
    \includegraphics[width=0.8\linewidth]{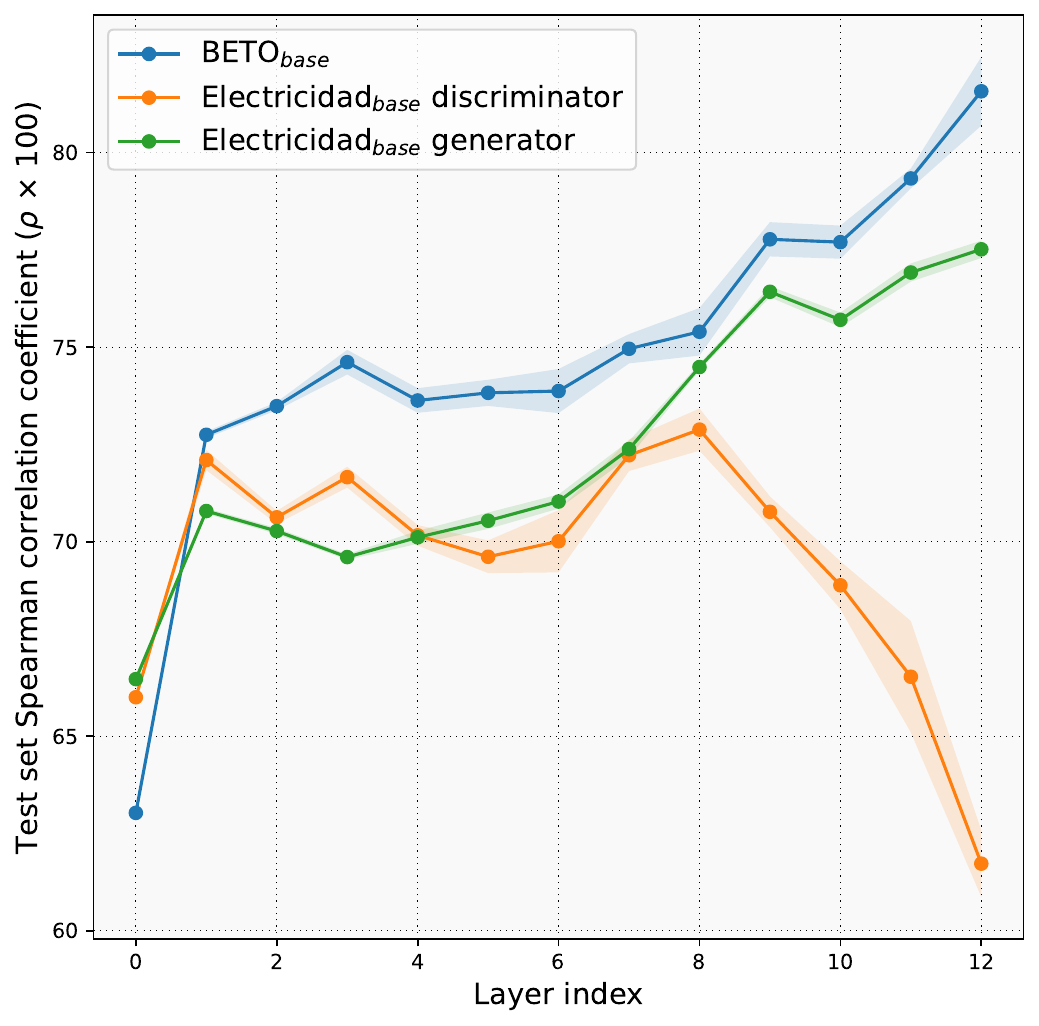}
    \caption{Test set Spearman correlation coefficients on Spanish STSB across all layers.}
    \label{fig:spanish}     
\end{figure}

To strengthen our findings, we provide experiments on three other languages: Korean, German, and Spanish. Due to the scarcity of pre-trained models and labeled data, we opt for these languages. All datasets used for the experiments were machine translated. The Korean STS \citep{ham-etal-2020-kornli} uses an internal neural machine translation engine for all training splits, although the validation set and test set were checked for errors by annotators. The other datasets used for training are completely automatically translated using the DeepL API \citep{huggingface:dataset:stsb_multi_mt}. The results for Korean, German, and Spanish are shown in Figures~\ref{fig:korean},~\ref{fig:german}, and~\ref{fig:spanish} respectively. 

The experiments for Korean (Figure~\ref{fig:korean}) show a performance drop for the KoELECTRA discriminator.\footnote{\url{https://huggingface.co/monologg/koelectra-base-v3-discriminator}} The final hidden state shows a Spearman correlation coefficient of $59.46$, while the ninth hidden state achieves $72.27$. The KoELECTRA generator\footnote{\url{https://huggingface.co/monologg/koelectra-base-v3-generator}} achieves a Spearman correlation coefficient of $78.50$ for the final hidden state, while for KLUE BERT\footnote{\url{https://huggingface.co/klue/bert-base}} it is $80.96$.

Experiments in German (Figure~\ref{fig:german}) exhibit a similar trend as in Korean and English. The GELECTRA discriminator\footnote{\url{https://huggingface.co/deepset/gelectra-base}} achieves a Spearman correlation coefficient of $56.22$, while for the seventh hidden state it is $71.15$. The difference between the GELECTRA generator\footnote{\url{https://huggingface.co/deepset/gelectra-base-generator}} and german BERT\footnote{\url{https://huggingface.co/google-bert/bert-base-german-cased}} is more apparent, with the highest Spearman correlation coefficients of $75.86$, and $79.54$, respectively.

Finally, the experiments in Spanish (Figure~\ref{fig:spanish}) are in line with the other languages as well. The Electricidad discriminator\footnote{\url{https://huggingface.co/mrm8488/electricidad-base-discriminator}} achieves a Spearman correlation coefficient of $61.73$ for the final hidden state. The highest value is $72.88$, which is achieved for the eight hidden state. A performance gap is present for the Electricidad generator\footnote{\url{https://huggingface.co/mrm8488/electricidad-base-generator}} and BETO model,\footnote{\url{https://huggingface.co/dccuchile/bert-base-spanish-wwm-cased}} scoring $77.51$, and $81.57$, respectively. To summarize, experiments in all languages exhibit a performance drop for the discriminator in the final layers.

\section{Performance for Various Model Sizes on STSB}
\label{appendix:C}

\begin{figure}[h!]
    \centering
    \includegraphics[width=0.8\linewidth]{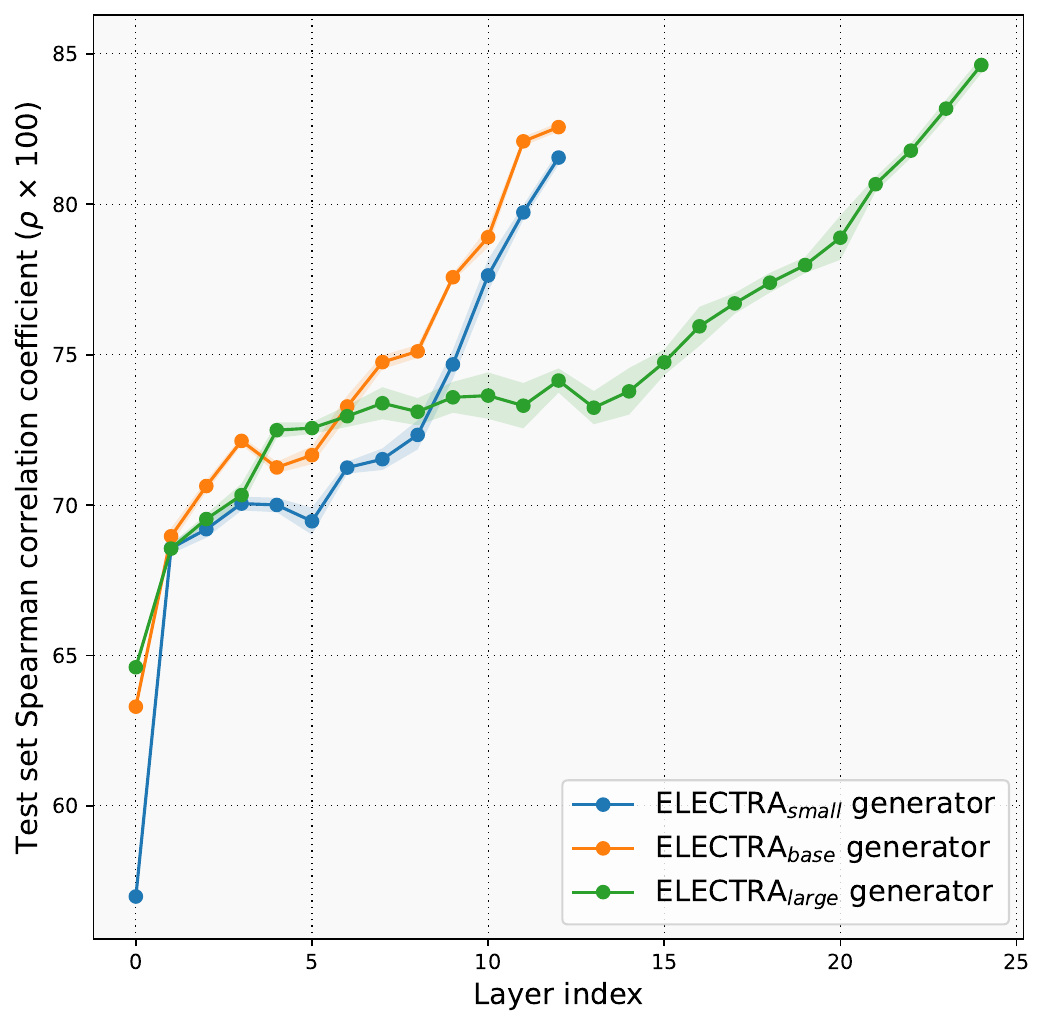}
    \caption{Test set Spearman correlation coefficients on STSB across various ELECTRA generator model sizes.}
    \label{fig:generator}
\end{figure}

\begin{figure}[h!]
    \centering
    \includegraphics[width=0.8\linewidth]{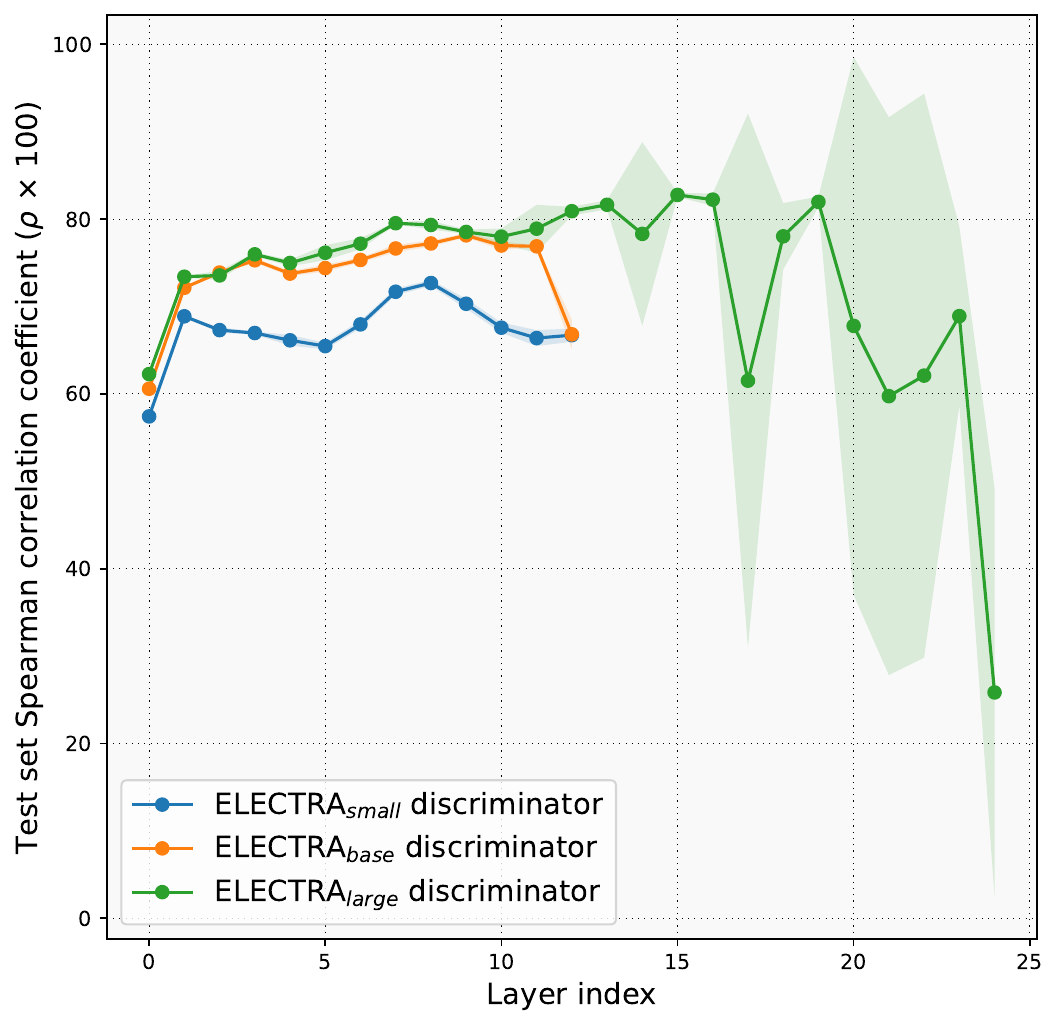}
    \caption{Test set Spearman correlation coefficients on STSB across various ELECTRA discriminator model sizes.}
    \label{fig:discriminator}
\end{figure}

\begin{figure}[h!]
    \centering
    \includegraphics[width=0.8\linewidth]{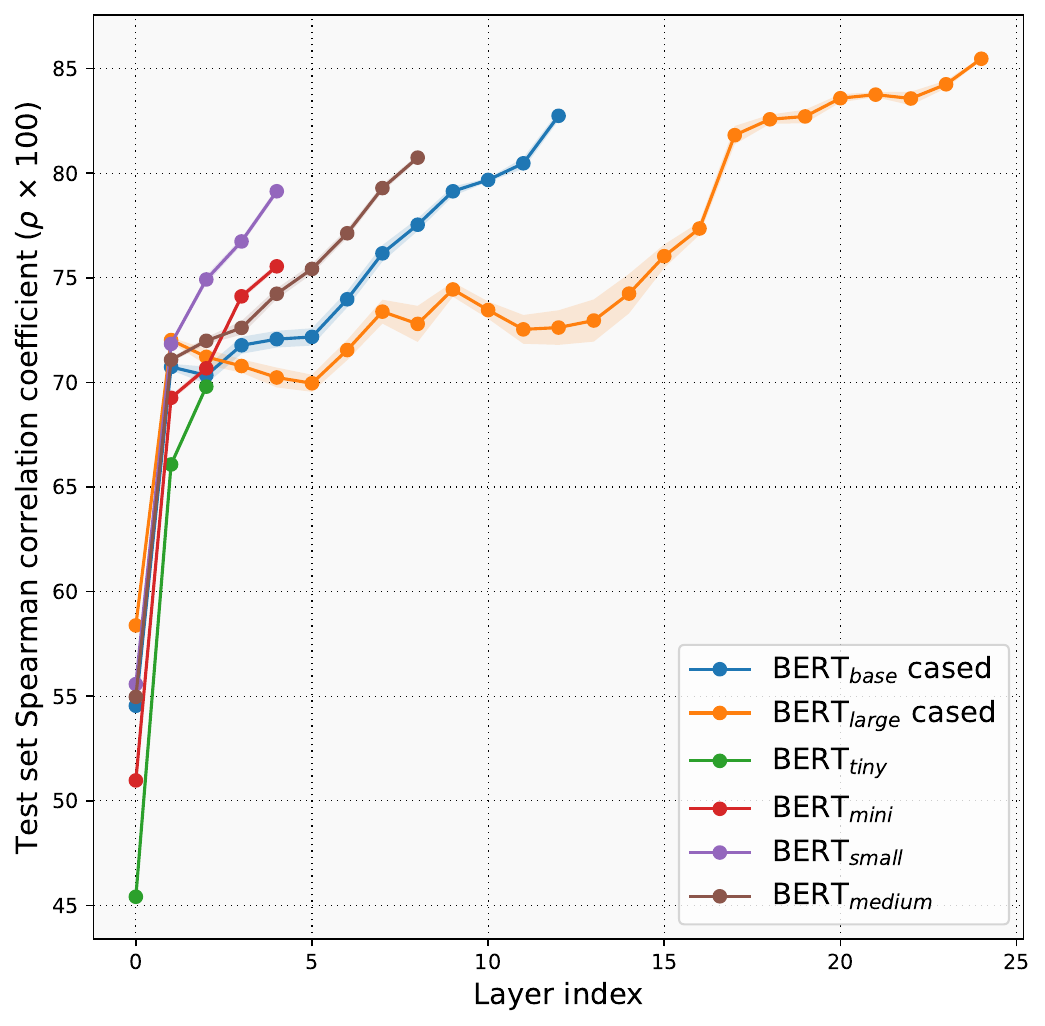}
    \caption{Test set Spearman correlation coefficients on STSB across various BERT model sizes.}
    \label{fig:bert}     
\end{figure}

We further strengthen our findings with experiments for various model sizes. The models used for the experiments are BERT, ELECTRA discriminator and ELECTRA generator. The figures~\ref{fig:generator},~\ref{fig:discriminator}, and ~\ref{fig:bert} show the test set Spearman correlation coefficients on STSB across various model sizes. The shaded are in the figures represents the standard deviation. 

For the ELECTRA generator experiments (Figure~\ref{fig:generator}), we use the small,\footnote{\url{https://huggingface.co/google/electra-small-generator}} base,\footnote{\url{https://huggingface.co/google/electra-base-generator}} and large\footnote{\url{https://huggingface.co/google/electra-large-generator}} sizes. The highest Spearman correlation coefficient scores are $81.55$, $82.57$, and $84.63$, respectively. All highest scores are achieved for the final hidden state, which is in line with other experiments.

The ELECTRA discriminator experiments (Figure~\ref{fig:discriminator}) show a performance drop present in the later layers, in agreement with previous results. The highest achieved Spearman correlation coefficients for the small,\footnote{\url{https://huggingface.co/google/electra-small-discriminator}} base,\footnote{\url{https://huggingface.co/google/electra-base-discriminator}} and large\footnote{\url{https://huggingface.co/google/electra-large-discriminator}} sizes are $72.68$, $78.14$, and $82.77$, achieved in layers $8$, $9$, and $15$, respectively.

For experiments with BERT (Figure~\ref{fig:bert}), there are more available standard sizes: tiny, mini, small, medium, base, and large. We apply our method to all of these sizes. The highest results are achieved for the last hidden state, while the scores for tiny, mini, small, medium, base, and large models are $69.80$, $75.55$, $79.13$, $80.74$, $82.74$, and $85.47$, respectively.

\section{Overview of Best-Performing Models Using TMFT on STSB}
\label{appendix:D}

Table~\ref{tab:pareto-table} represents the test set Spearman correlation coefficients and Pearson correlation coefficients on STSB for models that achieve the highest validation Spearman correlation coefficient using TMFT. The Table also includes the number of parameters of the truncated model. Bold values represent the highest values for the model family. BERT achieves the highest Spearman correlation coefficient for the last hidden layer, which is in line with other experiments. BERT$_\text{large}$ achieves the highest test set Spearman correlation coefficient of $85.47$ with $332.53$M parameters. For the ELECTRA discriminator small, base, and large, the hidden states with the highest validation Spearman correlation coefficient are $1$, $3$, and $12$, respectively. The ELECTRA discriminator greatly benefits from model truncation, which is demonstrated by the improvements of the Spearman correlation coefficient by $2.16$, $8.47$, and $55.06$ points, while using $8.69$M, $63.79$M, and $151.15$M parameters less for the small, base, and large models, respectively. Finally, the ELECTRA generator model family provides a parameter efficient alternative to BERT. The largest ELECTRA generator model uses only $50.74$M parameters and achieves a test set Spearman correlation coefficient of $84.63$, only $0.84$ points less than BERT$_\text{large}$, which has $281.79$M more parameters.

\begin{table}[h]
\centering
\resizebox{\columnwidth}{!}{
    \begin{tabular}{l|lr|cc}
        \toprule
        \textbf{Model} & \textbf{Layer} & \textbf{Params} & \textbf{Val}  & \textbf{Test}  \\
        \midrule
        BERT$_{\text{tiny}}$ & 2 & 4.37M & 78.20/77.57 & 69.80/70.64 \\
        BERT$_{\text{mini}}$ & 4 & 11.10M & 83.06/82.42 & 75.55/76.28 \\
        BERT$_{\text{small}}$ & 4 & 28.50M & 85.25/85.09 & 79.13/79.56 \\
        BERT$_{\text{medium}}$ & 8 & 41.11M & 85.74/85.46 & 80.74/81.02 \\
        BERT$_{\text{base}}$ & 12 & 107.72M & 86.07/85.98 & 82.74/83.03 \\
        BERT$_{\text{large}}$ & 24 & 332.53M & \textbf{88.33}/\textbf{88.31} & \textbf{85.47}/\textbf{85.68} \\
        \midrule
        ELECTRA$_{\text{D\:small}}$ & 1 & 4.76M & 79.74/79.27 & 68.88/69.64 \\
        ELECTRA$_{\text{D\:small\:last}}$ & 12 & 13.45M & 73.98/73.14 & 66.72/67.27 \\
        ELECTRA$_{\text{D\:base}}$ & 3 & 45.10M & 82.15/82.20 & 75.29/76.96 \\
        ELECTRA$_{\text{D\:base\:last}}$ & 12 & 108.89M & 72.41/71.62 & 66.82/67.23 \\
        ELECTRA$_{\text{D\:large}}$ & 12 & 182.94M & \textbf{84.74}/\textbf{84.88} & \textbf{80.90}/\textbf{81.15} \\
        ELECTRA$_{\text{D\:large\:last}}$ & 24 & 334.09M & 29.88/28.44 & 25.84/25.21 \\
        \midrule
        ELECTRA$_{\text{G\:small}}$ & 12 & 13.45M & 84.62/84.11 & 81.55/80.93 \\
        ELECTRA$_{\text{G\:base}}$ & 12 & 33.31M & 86.62/86.38 & 82.57/82.50 \\
        ELECTRA$_{\text{G\:large}}$ & 24 & 50.74M & \textbf{87.23}/\textbf{86.86} & \textbf{84.63}/\textbf{84.52} \\
        \bottomrule
    \end{tabular}
}
\caption{An overview of the test set Spearman correlation coefficients and Pearson correlation coefficients for various model families and model sizes. The reported test set values correspond to the model with the highest validation set Spearman correlation coefficient. For parameter calculation, the pooler layer is excluded as we do not use it.}
\label{tab:pareto-table}
\end{table}

\vfill

\section{TMFT on Randomly Initialized Models}
\label{appendix:E}

\begin{figure}[t!]
    \centering
    \includegraphics[width=0.8\linewidth]{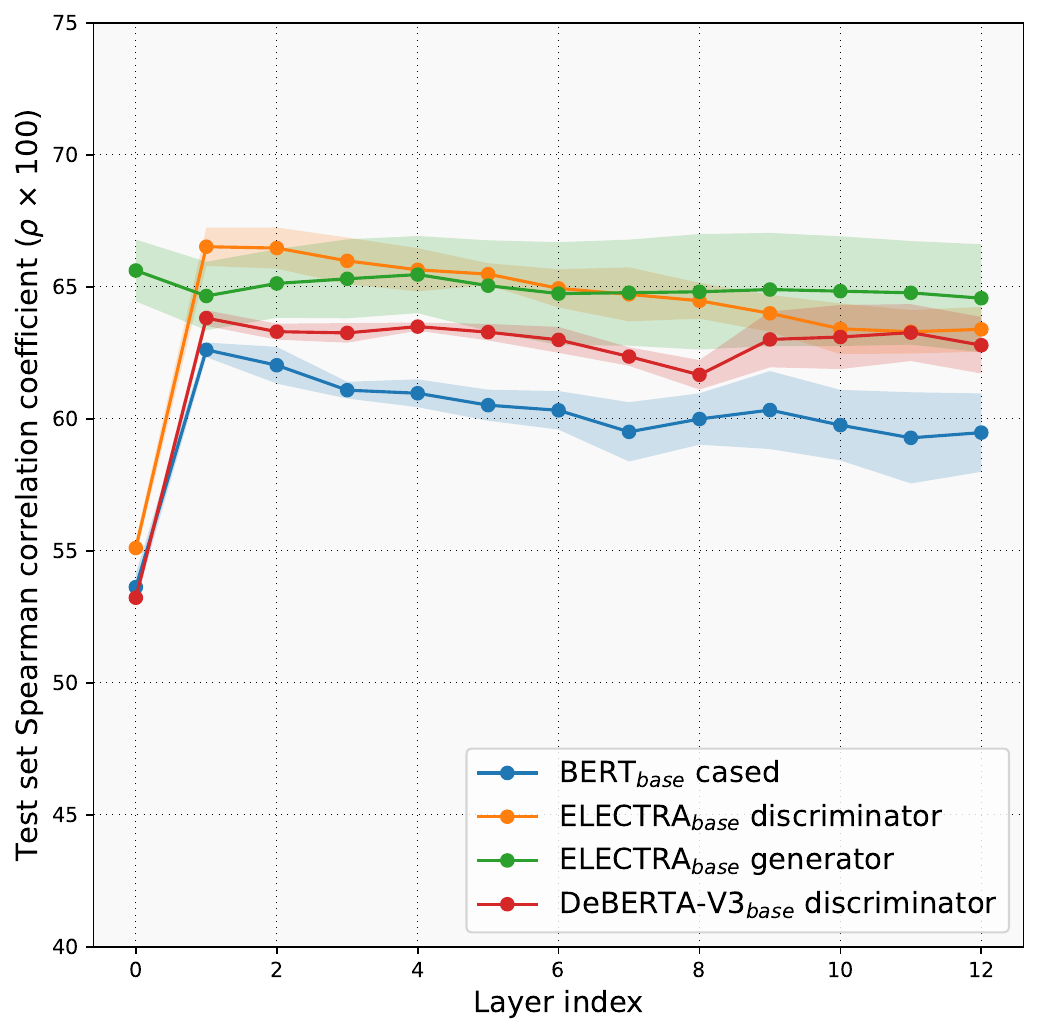}
    \caption{Test set Spearman correlation coefficients on STSB for randomly initialized models.}
    \label{fig:random}
\end{figure}

To verify whether excluding the pre-training step will shed more light on the performance drop, we provide an ablation study with randomly initialized models. Our hypothesis is that all models should roughly exhibit the same behaviour when fine-tuned up to a certain layer. Figure~\ref{fig:random} presents the result for TMFT on randomly initialized models. After layer zero, all models exhibit the same behaviour, with minor oscillations. The biggest difference is present in layer zero, where ELECTRA generator performs the best. The results suggest that architecture and tokenizer choice are not the cause of the performance drop. However, this does not exclude the effect of pre-training data, pre-training method, fine-tuning data, or hyperparameter choice.

\newpage

\section{Reproducibility}
\label{appendix:F}

We conducted our experiments on an AMD Ryzen Threadripper 3970X 32-Core Processor and a single RTX 3090 GPU with 24GB of RAM. Running the experiments took around 300 GPU hours. DAPT experiments with BERT and ELECTRA discriminator take around 7 GPU hours each, while for the ELECTRA generator it is around 4.5 GPU hours. Pre-training DeBERTaV3 took the longest, lasting around 30 hours. The word similarity fine-tuning takes up to 3 minutes, regardless of the model.

\end{document}